\documentclass[acmsmall,nonacm]{acmart}
\setcopyright{cc}
\copyrightyear{2023}
\acmYear{2023}
\acmDOI{}
\acmConference[XNLP23]{CMSC 848D: Explainable NLP}{2023}{College Park, MD}
\usepackage{todonotes}

\usepackage{caption}
\usepackage{subcaption}

\title{Towards Understanding In-Context Learning with Contrastive Demonstrations and Saliency Maps}

\author{Fuxiao Liu*, Paiheng Xu*, Zongxia Li*, Yue Feng, Hyemi Song\\ \quad University of Maryland, College Park}

\begin{document}
\maketitle


\section{abstract}
We investigate the role of various demonstration components in the in-context learning (ICL) performance of large language models (LLMs). Specifically, we explore the impacts of ground-truth labels, input distribution, and complementary explanations, particularly when these are altered or perturbed. We build on previous work, which offers mixed findings on how these elements influence ICL. To probe these questions, we employ explainable NLP (XNLP) methods and utilize saliency maps of contrastive demonstrations for both qualitative and quantitative analysis. Our findings reveal that flipping ground-truth labels significantly affects the saliency, though it's more noticeable in larger LLMs. Our analysis of the input distribution at a granular level reveals that changing sentiment-indicative terms in a sentiment analysis task to neutral ones does not have as substantial an impact as altering ground-truth labels. Finally, we find that the effectiveness of complementary explanations in boosting ICL performance is task-dependent, with limited benefits seen in sentiment analysis tasks compared to symbolic reasoning tasks. These insights are critical for understanding the functionality of LLMs and guiding the development of effective demonstrations, which is increasingly relevant in light of the growing use of LLMs in applications such as ChatGPT.

\section{Introduction}



Large language models (LLMs) show significant ability of in-context learning (ICL) for many NLP tasks \cite{zhao2023survey,yin2023survey,brown2020language}. ICL only requires a few input-label pairs for demonstrations and does not require fine-tuning on the model parameters. However, how each part of the demonstrations used in ICL drives the prediction remains an open research question. Previous works have mixed findings. For examples, although one might assume that ground-truth labels would have a similar impact on ICL as they do on supervised learning, \citet{min2022rethinking} finds that the ground truth input-label correspondence has little impact on the performance of end tasks. 
However, \citet{zhao2021calibrate} suggests that the example ordering has a strong impact.
More recently, \citet{wei2023larger} find that only LLMs with larger scales can learn the flipped input-label mapping. 

In this work, we use XNLP methods to understand which part of the demonstration contributes to the predictions more. 
We are interested in the impact of contrastive input-label demonstration pairs built in different ways, i.e., flipping the labels, changing the input, and adding complementary explanations as shown in Fig. \ref{fig:overview}. We then contrast the saliency maps of these contrastive demonstrations via qualitative and quantitative analysis.
Prior works \cite{min2022rethinking, wei2023larger, liu2020visual, brown2020language, liu2023aligning} show LLMs in relatively small scale, such as all GPT-3 models \cite{brown2020language} (based on categorization in \cite{wei2023larger}), cannot override prior knowledge from pretraining with demonstrations presented in-context, which means LLMs do not flip their predictions when the ground-truth labels are flipped in the demonstrations \cite{min2022rethinking}. 
However, \citet{wei2023larger} show larger models like InstructGPT \cite{ouyang2022training} (specifically the \texttt{text-davinci-002} checkpoint) and PaLM-540B \cite{chowdhery2022palm} have the emergent ability to override prior knowledge in the same setting.
We partly reproduce the results from previous work \cite{min2022rethinking, wei2023larger} on a sentiment classification task and find that the ground-truth labels in the demonstration are less salient after label flipping. 

Meanwhile, as the other important part of the demonstrations, the effect of input distribution is understudied. \citet{min2022rethinking} change the whole input to random words and \citet{wei2023larger} do no investigate input distribution at all. Therefore, we investigate the impact of input distribution at a fine-grained level, where we edit the input text's different components in correspondence to task-specific purposes. In the case of sentiment analysis, we change the sentiment-indicative terms in the input text of demonstrations to sentiment-neutral ones. We find that such input perturbation (neutralization) does not have as large impact as changing ground-truth label do. We suspect the models rely on pretrained knowledge to make fairly good predictions because the averaged importance scores for neutralized terms are smaller than the ones of original sentiment-indicative terms. 

Additionally, we find that complementary explanations do not necessarily benefit sentiment analysis task as they do for symbolic reasoning tasks as shown in \cite{ye2022complementary}, even though the saliency maps suggest the explanations tokens are as salient as the original input tokens. This suggests that we need to carefully generate complementary explanations and evaluate whether the target task would benefit from them when trying to boost ICL performance with such technique.


We hope the findings of this study can help researchers better understand the mechanism of LLMs and provide insights for practitioners when curating the demonstrations. Especially with the recent popularity of ChatGPT, we hope this study can help people from various domains have a better user experience with LLMs. The code for this study is available at \href{https://github.com/paihengxu/XICL}{https://github.com/paihengxu/XICL}.

\section{Background}

\subsection{Understanding ICL}
Large language models (LLMs) show significant ability of in-context learning (ICL) for many NLP tasks \cite{zhao2023survey,yin2023survey,brown2020language,liu2023documentclip}. \citet{min2022rethinking} show that presenting random ground truth labels in the demonstrations does not substantially affect performance. They also change other parts of the demonstrations (e.g., label space, distribution of the input text and overall sequence format) and find these factors are the key drivers for the end task performance. 
\citet{wei2023larger} concentrates on labels by comparing LMs across different size scales with two variants that have flipped labels or semantically-unrelated labels. They find that only large LMs can flip the predictions to follow flipped demonstrations. 
\citet{akyurek2022learning} and \citet{xie2021explanation} try to understand in-context learning by training transformer-based in-context learners on small-scale synthetic datasets.


\subsection{Saliency Maps}
\subsubsection{Gradient-based Methods} 
For models with parameter access, we can estimate the importance of an input token using derivative of output w.r.t that token. The most basic method assigns importance by the gradient \cite{simonyan2014deep}. However, it suffers from some known issues such as sensitivity to slight perturbations, saturated outputs, and discontinuous gradient. 
SmoothGrad \cite{smilkov2017smoothgrad} reduces the noise in the importance scores by adding Gaussian noise to the original input. Integrated Gradients (IG) \cite{pmlr-v70-sundararajan17a} computes a line integral of the vanilla saliency from a baseline point to the input in the feature space.

\subsubsection{Perturbation-based Methods}
An alternative approach to generating saliency maps using input perturbations can be applied to black-box models.
Instead, the process involves systematically altering the input data (i.e., words, phrases, and sentences) and observing the changes in the model's output. 
We plan to start with the standard method that falls into this category, LIME \cite{ribeiro2016should}.
The process involves creating perturbed versions of an input instance, passing them through the model, training a local linear model on the perturbed inputs and their corresponding predictions, and extracting feature importances from the local model. 


\section{Approach}

Despite previous efforts on understanding ICL \cite{min2022rethinking, wei2023larger, akyurek2022learning, xie2021explanation}, we are the first attempt to understand ICL using XNLP techniques to the best of our knowledge. We build contrastive demonstrations in various ways and contrast the saliency maps of these contrastive demonstrations to the ones of the original demonstrations to better understand ICL. 

\begin{figure}[t]
    \centering
      \includegraphics[width=1\textwidth]{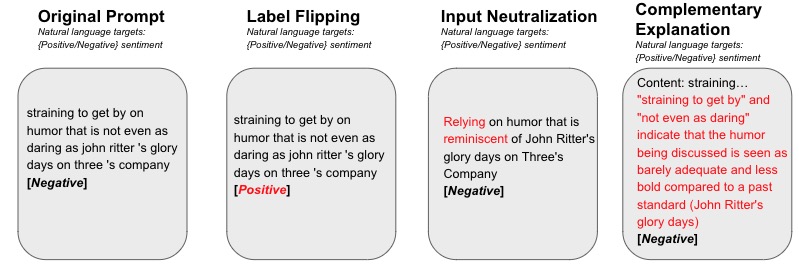}
    \caption{An overview of three ways to build contrastive demonstrations - flipping labels, perturbing (neutralizing) input, and adding complementary explanations. The contrastive parts are colored in red. 
    }
    \label{fig:overview}
\end{figure}

We adopt the following three methods to build contrastive demonstrations: flipping labels, perturbing (neutralizing) input, and adding complementary explanations, as shown in Fig. \ref{fig:overview}. We follow \cite{min2022rethinking, wei2023larger} to flip the labels in the demonstration. \citet{min2022rethinking} changed input text distribution to random English words, we focus on a task-specific perturbation in this study. Since we use sentiment analysis as our task and adjectives are strong and sometimes causal indicators of the prediction in this task, we neutralize adjectives in the demonstrations. \citet{ye2022complementary} show that adding complementary explanations benefits ICL. We want to investigate how important are these explanations using saliency map methods.
We hope that comparing the saliency maps of these contrastive prompts with the ones of the original prompt would give us insights into how different parts of the demonstrations contribute to ICL predictions.
 
\section{Experimental Set-up}
\paragraph{Dataset} We choose \textbf{SST-2} ~\cite{sst2013}, a sentiment analysis task, as our baseline task to explain ICL paradigm. Due to budget limitations and to follow \cite{min2022rethinking, wei2023larger}, we randomly sampled 288 examples that are not shorter than 20 tokens from the SST-2 training set as the test set. Additionally, we randomly sample 20 examples for generating saliency maps.

\subsection{Demonstration Selection}
\begin{figure}
     \centering
     \begin{subfigure}[b]{\textwidth}
         \centering
         \includegraphics[width=\textwidth]{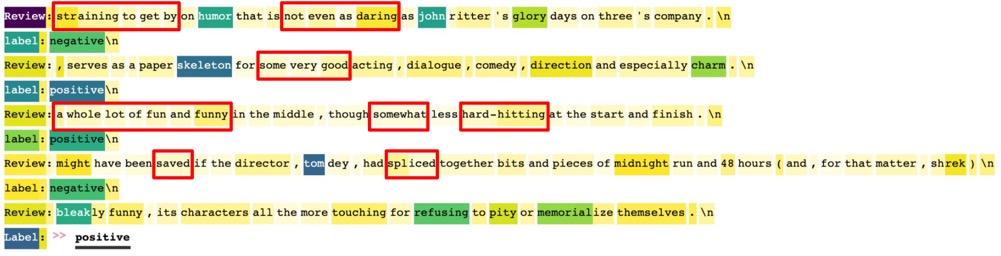}
         \caption{Original prompt}
         \label{fig:original}
     \end{subfigure}
     \par\bigskip
     \begin{subfigure}[b]{\textwidth}
         \centering
         \includegraphics[width=\textwidth]{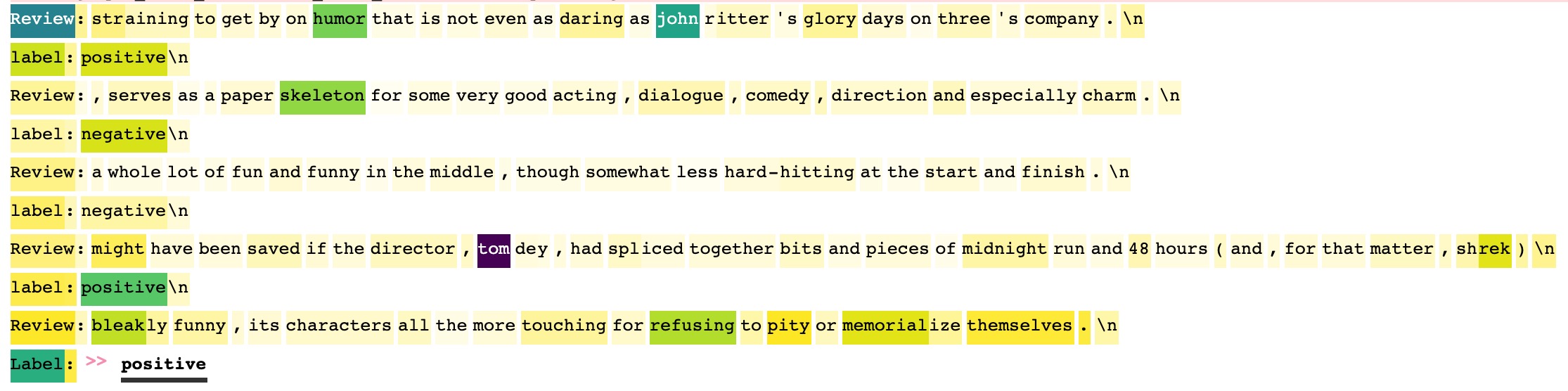}
         \caption{Prompt with label flipping in the demonstrations}
         \label{fig:label_flip}
     \end{subfigure}
     \par\bigskip
     \begin{subfigure}[b]{\textwidth}
         \centering
         \includegraphics[width=\textwidth]{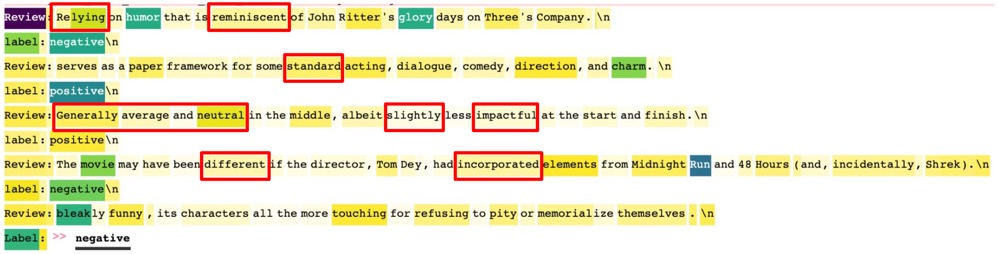}
         \caption{Prompt with input perturbation (neutralization) in the demonstrations}
         \label{fig:input_neutral}
     \end{subfigure}
     \par\bigskip
     \begin{subfigure}[b]{\textwidth}
         \centering
         \includegraphics[width=\textwidth]{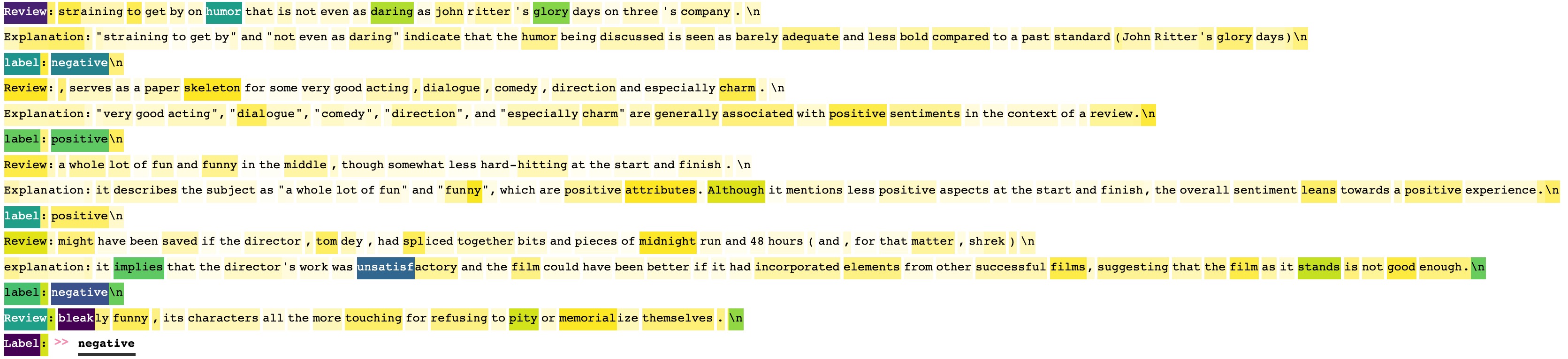}
         \caption{Prompt with complementary explanations in the demonstrations}
         \label{fig:explanation}
     \end{subfigure}
        \caption{Full prompts (demonstration + test example) used for original demonstration and three contrastive variants. Tokens are color-coded by saliency scores for GPT2 generated by IG. The red box in original and neutralized prompts indicates manually selected sentiment-indicative and sentiment-neutral terms that we used for saliency map comparison.}
        \label{fig:demo}
\end{figure}

Other than the 288 examples we picked from the training set, we also hand picked 4 example demonstrations to test language models' in-context-learning ability. We picked two examples with positive labels and two examples with negative labels to have an even distribution of classes in our demonstrations. The demonstrations have significant word indicators for positiveness and negativeness. As in Fig. ~\ref{fig:demo}, shows full demonstrations under four conditions: original demonstrations, label flipping, input neutralization, and adding explanation for each demonstration. 

\paragraph{Label Flipping}
We flip the binary label for each of the prompt and use them as the demonstrations during testing.

\paragraph{Input Neutralization}
For each of the demonstration, we prompted GPT-4 with its review, and asked it to replace indicative words or phrases (positive  or negative phrases) that could lead to the labels for the prompts with neutral words and phrases. After GPT-4 generates a perturbed version of the original review, we manually examine the validity of the perturbed prompts and make sure changes from the original prompts are minimal.

\paragraph{Complementary Explanation}
We add a complementary explanation for each of the review demonstration as shown in Fig ~\ref{fig:explanation}. The explanations are generated by prompting GPT-4 with \lq{Can you give an explanation to why (REVIEW) is labled positive/negative?}\rq. After each explanation was generated, we manually rephrase the explanation to make it shorter and more concise.

\subsection{Baseline LMs and Metric}
We first evaluate accuracy of the following models on the sampled SST-2 dataset.
\paragraph{Fine-tuned BERT}
We used BERT fine-tuned on SST-2 dataset to get a supervised accuracy of a language model\footnote{distilbert-base-uncased-finetuned-sst-2-english} for a better comparison on other models' accuracy on the same dataset without fine-tuning. 

\paragraph{ChatGPT-3.5-turbo}
We used the openAI gpt-3.5-turbo, with maximum number response token length set to 4096 tokens, temperature set to 0 (with no randomness of the outputs), and top\_n set to 1, which always gives the top 1 prediction of the model. 

\paragraph{Instruct-GPT}
We used \texttt{text-davinci-002}, a member of the Instruct-GPT family, with temperature 0, maximum tokens generated equal to 50 (to keep cost in control), and top\_n also set to 1.

\paragraph{GPT-2}
We select GPT-2 \footnote{https://huggingface.co/gpt2} with 124M parameters, the smallest version of GPT-2. The GPT-2 model does not sample when generating the output (i.e., temperature set to 0).
We are not able to run larger GPT-2 models like GPT-medium, GPT-large, and GPT-XL on a single GPU available at hand.

\paragraph{Metric}
We use the accuracy to evaluate sentiment classification. We also use T-test to verify our hypothesis on the saliency map patterns for the three contrastive demonstrations.

\subsection{Saliency Map Methods}
\label{sec:saliency_methods}
For gradient-based method, we use IG \cite{pmlr-v70-sundararajan17a} for models with parameter access (i.e., GPT-2) with the implementation from Ecco \cite{alammar2021ecco}. For black-box models (i.e., \texttt{text-davinci-002} from Instruct-GPT family), we use LIME\footnote{https://github.com/marcotcr/lime} to explain the Instruct-GPT classifier. 
We use the LimeTextExplainer to explain the instance, where the num\_features is $20$ and the number of neighbors is $5$.
We choose such a setting because of the budget limit as it requires fewer perturbations and interactions with Instruct-GPT API. It generates more ``sparse'' saliency maps, we discuss this in detail in Section \ref{sec:results_saliency}.
The hyperparameters for GPT-2 and GPT-3 and prompts are the same ones we used for the accuracy evaluation.  
We only generated saliency maps for GPT-2 and GPT-3 models due to time and compute constraints, future work could explore more models such as ChatGPT. 
 
\section{Findings}

\begin{figure*}[t]
    \centering
      \includegraphics[width=0.6\textwidth]{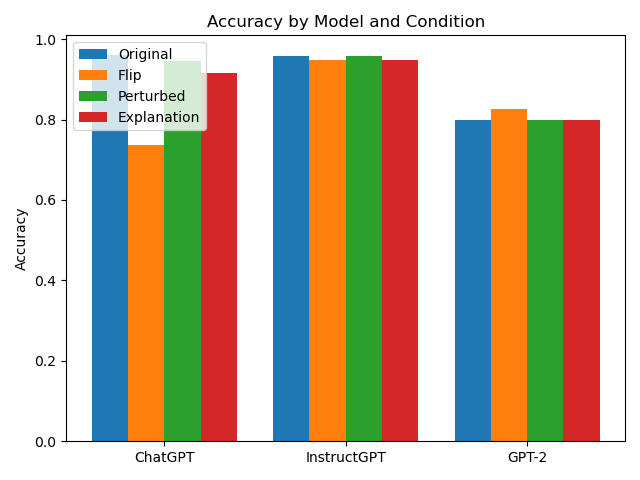}
    \caption{Model Performance under the four conditions, with \textbf{four} demonstrations given}
    \vspace{-0.1in}
    \label{fig:4demos}
\end{figure*}



\subsection{Prediction Performance of the Three Contrastive Demonstrations}
\begin{figure*}[t]
    \centering
      \includegraphics[width=0.6\textwidth]{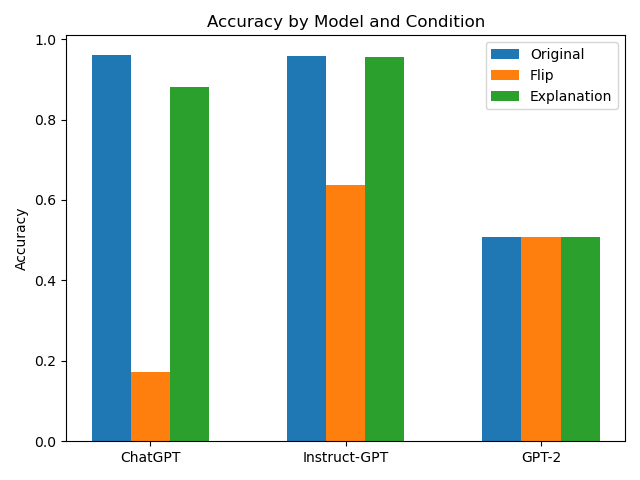}
    \caption{Model Performance under the four conditions, with \textbf{eight} demonstrations given.}
    \vspace{-0.1in}
    \label{fig:8demos}
\end{figure*}

We tested the performance of GPT-3.5-Turbo, InstructGPT, and GPT-2 on the 288 selected test examples with four types of demonstrations, i.e., original, label flipping, input neutralization, and complementary explanations. The results are shown in Fig. ~\ref{fig:4demos} and Fig. ~\ref{fig:8demos} \footnote{We excluded input perturbation from Fig. \ref{fig:8demos} because we perturbed the demonstrations differently while testing for 8 demonstrations (replace indicative words or phrases with opposite meaning words and phrases) and cannot afford to rerun the experiments in terms of time and cost.}. 
We find that for label-flipping demonstration, ChatGPT-Turbo-3.5 leads to the greatest degradation in performance, 
from accuracy 96\% to 73\% when given 4 demonstrations and to $17\%$ when given 8 demonstrations. The performance of InstructGPT drops by a smaller amount in both scenarios.
This is consistent with the findings from \cite{wei2023larger, min2022rethinking}, i.e., large LMs' (Instruct-GPT and ChatGPT) performance drops with increased number of the labels flipped in exemplars.
Although the model parameter sizes are the same between GPT-3.5-Turbo and InstructGPT, GPT-3.5-Turbo seems to have a much stronger in-context-learning ability than InstructGPT. 

We also noticed that GPT-2 model has much lower performance when given 4 demonstrations and almost predicts negative for all test examples when given 8 demonstrations. It is thus relatively insensitive to different types of contrastive demonstrations. The fact that the performance change of label-flipping is much smaller for GPT-2 compared with ChatGPT and Instruct-GPT also verifies the findings of emergent overriding ability of large LMS from \cite{wei2023larger}. 

On the other hand, we observe much smaller impacts for input neutralization and explanation. The small impact of input neutralization may be due to the fact that LMs are still able to make predictions using pretrained knowledge, especially for a relatively ``easy'' task of sentiment analysis. Adding complementary explanation to each examplar does not benefit the model performance for both four-demonstration and eight-demonstration scenarios. One of the reasons might be that the task chosen is too trivial for explanations to be useful.

These mixed results further inspire us to contrast the patterns of the saliency maps generated by smaller LLMs and large LLMs as all the language models tested are built upon transformer architecture.

\subsection{Comparison of the Saliency Maps}
\label{sec:results_saliency}

Due to the GPT-2's poor performance and compute cost when given 8 demonstrations, we use the setting of 4 demonstrations for saliency map comparisons as shown in Fig. \ref{fig:demo} and Fig. \ref{fig:demo_lime}. 

\subsubsection{Label Flipping}

\paragraph{Hypothesis} The labels in the demonstration are less important after model flipping for smaller LMs (GPT2) but more important for large LMs (\texttt{text-davinci-002 from Instruct-GPT}).
\paragraph{Analysis} For example as in Fig. \ref{fig:original} and Fig. \ref{fig:label_flip}, the importance of the output label in the demonstration decreases from the original prompt to the label-flipped one. This suggests that the model might pay less attention to the flipped label due to its inconsistency with the input, which results in insensitivity to label flipping in the demonstrations. We expect smaller LMs (GPT2) and large LMs (\texttt{text-davinci-002} from Instruct-GPT) to have different behaviors because \citet{wei2023larger} show only large LMs have the ability to override prior knowledge from pertaining to the one from demonstrations, which is also supported by our results from Fig. \ref{fig:4demos} and Fig. \ref{fig:8demos}. 

For GPT2, on average, $3.35 / 4$ of the labels in the demonstration have decreased saliency scores when the demo labels are flipped. Moreover, the average saliency scores of the $4$ demo labels \textbf{decrease} for all $20$ test examples. The p-value from a T-test for comparing average saliency scores ($N=20$) between original and label-flipped demonstrations is $< 0.001$.
For InstructGPT, the average saliency scores \textbf{increase} for $16/20$ test examples with a p-value of $0.23$ from a similar T-test as above (Fig. \ref{fig:LIME2}). As InstructGPT achieves around $60\%$ accuracy in Fig. \ref{fig:8demos}, we expect Instruct-GPT (with 8 demonstrations) and ChatGPT to have a more significant result as it shows the ability to fully override prior pretrained knowledge.

\subsubsection{Input Perturbation (Neutralization)}

\paragraph{Hypothesis} The sentiment-indicative terms in the original prompt are more important than sentiment-neutral terms in the neutralized prompt.

\paragraph{Analysis} The hypothesis is derived from the definition and our intuition of the sentiment analysis task. Sentiment-indicative terms are important to make sentiment predictions. To validate this hypothesis, we contrast the original and neutralized prompts and manually pick different tokens with sentiment orientations. The selected tokens are highlighted in Fig. \ref{fig:original} and Fig. \ref{fig:input_neutral} with red boxes respectively. We then compute the average saliency scores for each of the 20 test examples.

We find that, for GPT2, the average saliency scores for sentiment-indicative terms in the original prompt are higher than their contrastive parts in the neutralized prompt for all 20 test examples with a p-value of $<0.001$ from a T-test. However, for Instruct-GPT, we find that the sentiment-indicative terms in the original prompt are equal or higher in $9/20$ test examples with a p-value of $0.17$ from a similar T-test as above. We note that, as mentioned in Section \ref{sec:saliency_methods}, the saliency maps for Instruct-GPT generated by LIME are sparse and have a lot of zeros as shown in Fig. \ref{fig:demo_lime}. This may lead to a mixed result with a less significant T-test result.

\subsubsection{Complementary Explanation}
Previous work \citet{ye2022complementary} shows complementary explanations are beneficial for symbolic reasoning tasks including Letter Concatenation, Coin Flips, and Grade School Math. However, as we show in Fig. \ref{fig:4demos}, complementary explanations do not necessarily improve the performance for sentiment analysis which is considered as an ``easier'' task for LMs. From saliency maps for GPT2, we find that $16/20$ test examples have larger averaged saliency scores for tokens in the explanations than the ones in the review. On average, the averaged saliency scores for review tokens are $90\%$ of the ones for explanation tokens. So the gaps are small and explanations are just as important as the original review in this sense.  
We suspect the impact of complementary explanations varies based on the tasks at hand. It may benefit tasks requiring more logical reasoning but such conclusion requires more systematical evaluation on more benchmark datasets and we leave it to future work.

\section{Limitations}

One potential limitation of our work is that we only select 288 samples, only 20 examples for the saliency map and only 5 neighbors in the LIME model. This is because of budge limitation and Openai daily request limit. In the future, a larger dataset is needed for the evaluation. Additionally, all the demos we used for the ICL is randomly selected, which will influence the model's accuracy performance in some cases. Therefore, a better method would be using different methods to select the demos, like similar instance retrieval. Finally, we only pick the saliency map as the explanation method. In the future, more explanation models will support our conclusion better. In addition, since we find out that adding explanations to the demonstrations leads to performance degradation, we would like to examine the role of explanation on other tasks, or with more examples being tested for the SST-2 dataset. However, due to time-constrains and budget issues, we are unable to conduct the experiment on more complex tasks, such as grade school math problems, or common sense reasoning to further examine why adding explanation leads to performance degradation on sentiment analysis task. 

\section{Conclusion}


In this study, we used XNLP techniques to study how ICL works by investigating the performance of contrastive input-label demonstration pairs built in different ways, i.e., flipping the labels, changing the input, and adding complementary explanations, and contrasting their saliency maps with the original demonstration via qualitative and quantitative analysis.
We partly reproduced the results from previous work \cite{min2022rethinking, wei2023larger} on a sentiment classification task and found that the ground-truth labels in the demonstration are less salient after label flipping. We also found that neutralizing the sentiment-indicative terms in the input does not have as large impact as changing ground-truth label do. The models may rely on pretrained knowledge to make fairly good predictions because the averaged importance scores for neutralized terms are smaller than the ones of original sentiment-indicative terms. Additionally, complementary explanations do not necessarily benefit sentiment analysis task as they do for symbolic reasoning tasks as shown in \cite{ye2022complementary}. We hope the findings of this study can help researchers better understand the mechanism of LLMs and provide insights for practitioners when curating the demonstrations.

Future works can experiment with more LMs and more benchmark tasks and datasets to verify the findings of this study and make them more generalizable. Moreover, this work only focused on analyzing the saliency maps of demonstrations. Future work can investigate how demonstrations interact with the query test examples by adjusting exemplars in the demonstration based on their semantic distance from the query examples.
It is also interesting to compare gradient-based saliency map methods with perturbation-based ones and see how the saliency maps differ with the same input and model.

\bibliographystyle{ACM-Reference-Format}
\bibliography{sample-base}


\begin{thebibliography}{20}


\ifx \showCODEN    \undefined \def \showCODEN     #1{\unskip}     \fi
\ifx \showDOI      \undefined \def \showDOI       #1{#1}\fi
\ifx \showISBNx    \undefined \def \showISBNx     #1{\unskip}     \fi
\ifx \showISBNxiii \undefined \def \showISBNxiii  #1{\unskip}     \fi
\ifx \showISSN     \undefined \def \showISSN      #1{\unskip}     \fi
\ifx \showLCCN     \undefined \def \showLCCN      #1{\unskip}     \fi
\ifx \shownote     \undefined \def \shownote      #1{#1}          \fi
\ifx \showarticletitle \undefined \def \showarticletitle #1{#1}   \fi
\ifx \showURL      \undefined \def \showURL       {\relax}        \fi
\providecommand\bibfield[2]{#2}
\providecommand\bibinfo[2]{#2}
\providecommand\natexlab[1]{#1}
\providecommand\showeprint[2][]{arXiv:#2}

\bibitem[Aky{\"u}rek et~al\mbox{.}(2022)]%
        {akyurek2022learning}
\bibfield{author}{\bibinfo{person}{Ekin Aky{\"u}rek}, \bibinfo{person}{Dale Schuurmans}, \bibinfo{person}{Jacob Andreas}, \bibinfo{person}{Tengyu Ma}, {and} \bibinfo{person}{Denny Zhou}.} \bibinfo{year}{2022}\natexlab{}.
\newblock \showarticletitle{What learning algorithm is in-context learning? investigations with linear models}.
\newblock \bibinfo{journal}{\emph{arXiv preprint arXiv:2211.15661}} (\bibinfo{year}{2022}).
\newblock


\bibitem[Alammar(2021)]%
        {alammar2021ecco}
\bibfield{author}{\bibinfo{person}{J Alammar}.} \bibinfo{year}{2021}\natexlab{}.
\newblock \showarticletitle{Ecco: An open source library for the explainability of transformer language models}. In \bibinfo{booktitle}{\emph{Proceedings of the 59th Annual Meeting of the Association for Computational Linguistics and the 11th International Joint Conference on Natural Language Processing: System Demonstrations}}. \bibinfo{pages}{249--257}.
\newblock


\bibitem[Brown et~al\mbox{.}(2020)]%
        {brown2020language}
\bibfield{author}{\bibinfo{person}{Tom Brown}, \bibinfo{person}{Benjamin Mann}, \bibinfo{person}{Nick Ryder}, \bibinfo{person}{Melanie Subbiah}, \bibinfo{person}{Jared~D Kaplan}, \bibinfo{person}{Prafulla Dhariwal}, \bibinfo{person}{Arvind Neelakantan}, \bibinfo{person}{Pranav Shyam}, \bibinfo{person}{Girish Sastry}, \bibinfo{person}{Amanda Askell}, {et~al\mbox{.}}} \bibinfo{year}{2020}\natexlab{}.
\newblock \showarticletitle{Language models are few-shot learners}.
\newblock \bibinfo{journal}{\emph{Advances in neural information processing systems}}  \bibinfo{volume}{33} (\bibinfo{year}{2020}), \bibinfo{pages}{1877--1901}.
\newblock


\bibitem[Chowdhery et~al\mbox{.}(2022)]%
        {chowdhery2022palm}
\bibfield{author}{\bibinfo{person}{Aakanksha Chowdhery}, \bibinfo{person}{Sharan Narang}, \bibinfo{person}{Jacob Devlin}, \bibinfo{person}{Maarten Bosma}, \bibinfo{person}{Gaurav Mishra}, \bibinfo{person}{Adam Roberts}, \bibinfo{person}{Paul Barham}, \bibinfo{person}{Hyung~Won Chung}, \bibinfo{person}{Charles Sutton}, \bibinfo{person}{Sebastian Gehrmann}, {et~al\mbox{.}}} \bibinfo{year}{2022}\natexlab{}.
\newblock \showarticletitle{Palm: Scaling language modeling with pathways}.
\newblock \bibinfo{journal}{\emph{arXiv preprint arXiv:2204.02311}} (\bibinfo{year}{2022}).
\newblock


\bibitem[Liu et~al\mbox{.}(2023a)]%
        {liu2023aligning}
\bibfield{author}{\bibinfo{person}{Fuxiao Liu}, \bibinfo{person}{Kevin Lin}, \bibinfo{person}{Linjie Li}, \bibinfo{person}{Jianfeng Wang}, \bibinfo{person}{Yaser Yacoob}, {and} \bibinfo{person}{Lijuan Wang}.} \bibinfo{year}{2023}\natexlab{a}.
\newblock \showarticletitle{Aligning Large Multi-Modal Model with Robust Instruction Tuning}.
\newblock \bibinfo{journal}{\emph{arXiv preprint arXiv:2306.14565}} (\bibinfo{year}{2023}).
\newblock


\bibitem[Liu et~al\mbox{.}(2023b)]%
        {liu2023documentclip}
\bibfield{author}{\bibinfo{person}{Fuxiao Liu}, \bibinfo{person}{Hao Tan}, {and} \bibinfo{person}{Chris Tensmeyer}.} \bibinfo{year}{2023}\natexlab{b}.
\newblock \showarticletitle{DocumentCLIP: Linking Figures and Main Body Text in Reflowed Documents}.
\newblock \bibinfo{journal}{\emph{arXiv preprint arXiv:2306.06306}} (\bibinfo{year}{2023}).
\newblock


\bibitem[Liu et~al\mbox{.}(2020)]%
        {liu2020visual}
\bibfield{author}{\bibinfo{person}{Fuxiao Liu}, \bibinfo{person}{Yinghan Wang}, \bibinfo{person}{Tianlu Wang}, {and} \bibinfo{person}{Vicente Ordonez}.} \bibinfo{year}{2020}\natexlab{}.
\newblock \showarticletitle{Visual news: Benchmark and challenges in news image captioning}.
\newblock \bibinfo{journal}{\emph{arXiv preprint arXiv:2010.03743}} (\bibinfo{year}{2020}).
\newblock


\bibitem[Min et~al\mbox{.}(2022)]%
        {min2022rethinking}
\bibfield{author}{\bibinfo{person}{Sewon Min}, \bibinfo{person}{Xinxi Lyu}, \bibinfo{person}{Ari Holtzman}, \bibinfo{person}{Mikel Artetxe}, \bibinfo{person}{Mike Lewis}, \bibinfo{person}{Hannaneh Hajishirzi}, {and} \bibinfo{person}{Luke Zettlemoyer}.} \bibinfo{year}{2022}\natexlab{}.
\newblock \showarticletitle{Rethinking the Role of Demonstrations: What Makes In-Context Learning Work?}
\newblock \bibinfo{journal}{\emph{arXiv preprint arXiv:2202.12837}} (\bibinfo{year}{2022}).
\newblock


\bibitem[Ouyang et~al\mbox{.}(2022)]%
        {ouyang2022training}
\bibfield{author}{\bibinfo{person}{Long Ouyang}, \bibinfo{person}{Jeffrey Wu}, \bibinfo{person}{Xu Jiang}, \bibinfo{person}{Diogo Almeida}, \bibinfo{person}{Carroll Wainwright}, \bibinfo{person}{Pamela Mishkin}, \bibinfo{person}{Chong Zhang}, \bibinfo{person}{Sandhini Agarwal}, \bibinfo{person}{Katarina Slama}, \bibinfo{person}{Alex Ray}, {et~al\mbox{.}}} \bibinfo{year}{2022}\natexlab{}.
\newblock \showarticletitle{Training language models to follow instructions with human feedback}.
\newblock \bibinfo{journal}{\emph{Advances in Neural Information Processing Systems}}  \bibinfo{volume}{35} (\bibinfo{year}{2022}), \bibinfo{pages}{27730--27744}.
\newblock


\bibitem[Ribeiro et~al\mbox{.}(2016)]%
        {ribeiro2016should}
\bibfield{author}{\bibinfo{person}{Marco~Tulio Ribeiro}, \bibinfo{person}{Sameer Singh}, {and} \bibinfo{person}{Carlos Guestrin}.} \bibinfo{year}{2016}\natexlab{}.
\newblock \showarticletitle{" Why should i trust you?" Explaining the predictions of any classifier}. In \bibinfo{booktitle}{\emph{Proceedings of the 22nd ACM SIGKDD international conference on knowledge discovery and data mining}}. \bibinfo{pages}{1135--1144}.
\newblock


\bibitem[Simonyan et~al\mbox{.}(2014)]%
        {simonyan2014deep}
\bibfield{author}{\bibinfo{person}{K Simonyan}, \bibinfo{person}{A Vedaldi}, {and} \bibinfo{person}{A Zisserman}.} \bibinfo{year}{2014}\natexlab{}.
\newblock \showarticletitle{Deep inside convolutional networks: visualising image classification models and saliency maps}. In \bibinfo{booktitle}{\emph{Proceedings of the International Conference on Learning Representations (ICLR)}}. ICLR.
\newblock


\bibitem[Smilkov et~al\mbox{.}(2017)]%
        {smilkov2017smoothgrad}
\bibfield{author}{\bibinfo{person}{Daniel Smilkov}, \bibinfo{person}{Nikhil Thorat}, \bibinfo{person}{Been Kim}, \bibinfo{person}{Fernanda Vi{\'e}gas}, {and} \bibinfo{person}{Martin Wattenberg}.} \bibinfo{year}{2017}\natexlab{}.
\newblock \showarticletitle{Smoothgrad: removing noise by adding noise}.
\newblock \bibinfo{journal}{\emph{arXiv preprint arXiv:1706.03825}} (\bibinfo{year}{2017}).
\newblock


\bibitem[Socher et~al\mbox{.}(2013)]%
        {sst2013}
\bibfield{author}{\bibinfo{person}{Richard Socher}, \bibinfo{person}{Alex Perelygin}, \bibinfo{person}{Jean Wu}, \bibinfo{person}{Jason Chuang}, \bibinfo{person}{Christopher~D. Manning}, \bibinfo{person}{Andrew Ng}, {and} \bibinfo{person}{Christopher Potts}.} \bibinfo{year}{2013}\natexlab{}.
\newblock \showarticletitle{Recursive Deep Models for Semantic Compositionality Over a Sentiment Treebank}. In \bibinfo{booktitle}{\emph{Proceedings of the 2013 Conference on Empirical Methods in Natural Language Processing}}. \bibinfo{publisher}{Association for Computational Linguistics}, \bibinfo{address}{Seattle, Washington, USA}, \bibinfo{pages}{1631--1642}.
\newblock
\urldef\tempurl%
\url{https://aclanthology.org/D13-1170}
\showURL{%
\tempurl}


\bibitem[Sundararajan et~al\mbox{.}(2017)]%
        {pmlr-v70-sundararajan17a}
\bibfield{author}{\bibinfo{person}{Mukund Sundararajan}, \bibinfo{person}{Ankur Taly}, {and} \bibinfo{person}{Qiqi Yan}.} \bibinfo{year}{2017}\natexlab{}.
\newblock \showarticletitle{Axiomatic Attribution for Deep Networks}. In \bibinfo{booktitle}{\emph{Proceedings of the 34th International Conference on Machine Learning}} \emph{(\bibinfo{series}{Proceedings of Machine Learning Research}, Vol.~\bibinfo{volume}{70})}, \bibfield{editor}{\bibinfo{person}{Doina Precup} {and} \bibinfo{person}{Yee~Whye Teh}} (Eds.). \bibinfo{publisher}{PMLR}, \bibinfo{pages}{3319--3328}.
\newblock
\urldef\tempurl%
\url{https://proceedings.mlr.press/v70/sundararajan17a.html}
\showURL{%
\tempurl}


\bibitem[Wei et~al\mbox{.}(2023)]%
        {wei2023larger}
\bibfield{author}{\bibinfo{person}{Jerry Wei}, \bibinfo{person}{Jason Wei}, \bibinfo{person}{Yi Tay}, \bibinfo{person}{Dustin Tran}, \bibinfo{person}{Albert Webson}, \bibinfo{person}{Yifeng Lu}, \bibinfo{person}{Xinyun Chen}, \bibinfo{person}{Hanxiao Liu}, \bibinfo{person}{Da Huang}, \bibinfo{person}{Denny Zhou}, {et~al\mbox{.}}} \bibinfo{year}{2023}\natexlab{}.
\newblock \showarticletitle{Larger language models do in-context learning differently}.
\newblock \bibinfo{journal}{\emph{arXiv preprint arXiv:2303.03846}} (\bibinfo{year}{2023}).
\newblock


\bibitem[Xie et~al\mbox{.}(2021)]%
        {xie2021explanation}
\bibfield{author}{\bibinfo{person}{Sang~Michael Xie}, \bibinfo{person}{Aditi Raghunathan}, \bibinfo{person}{Percy Liang}, {and} \bibinfo{person}{Tengyu Ma}.} \bibinfo{year}{2021}\natexlab{}.
\newblock \showarticletitle{An explanation of in-context learning as implicit bayesian inference}.
\newblock \bibinfo{journal}{\emph{arXiv preprint arXiv:2111.02080}} (\bibinfo{year}{2021}).
\newblock


\bibitem[Ye et~al\mbox{.}(2022)]%
        {ye2022complementary}
\bibfield{author}{\bibinfo{person}{Xi Ye}, \bibinfo{person}{Srinivasan Iyer}, \bibinfo{person}{Asli Celikyilmaz}, \bibinfo{person}{Ves Stoyanov}, \bibinfo{person}{Greg Durrett}, {and} \bibinfo{person}{Ramakanth Pasunuru}.} \bibinfo{year}{2022}\natexlab{}.
\newblock \showarticletitle{Complementary Explanations for Effective In-Context Learning}.
\newblock \bibinfo{journal}{\emph{arXiv preprint arXiv:2211.13892}} (\bibinfo{year}{2022}).
\newblock


\bibitem[Yin et~al\mbox{.}(2023)]%
        {yin2023survey}
\bibfield{author}{\bibinfo{person}{Shukang Yin}, \bibinfo{person}{Chaoyou Fu}, \bibinfo{person}{Sirui Zhao}, \bibinfo{person}{Ke Li}, \bibinfo{person}{Xing Sun}, \bibinfo{person}{Tong Xu}, {and} \bibinfo{person}{Enhong Chen}.} \bibinfo{year}{2023}\natexlab{}.
\newblock \showarticletitle{A Survey on Multimodal Large Language Models}.
\newblock \bibinfo{journal}{\emph{arXiv preprint arXiv:2306.13549}} (\bibinfo{year}{2023}).
\newblock


\bibitem[Zhao et~al\mbox{.}(2023)]%
        {zhao2023survey}
\bibfield{author}{\bibinfo{person}{Wayne~Xin Zhao}, \bibinfo{person}{Kun Zhou}, \bibinfo{person}{Junyi Li}, \bibinfo{person}{Tianyi Tang}, \bibinfo{person}{Xiaolei Wang}, \bibinfo{person}{Yupeng Hou}, \bibinfo{person}{Yingqian Min}, \bibinfo{person}{Beichen Zhang}, \bibinfo{person}{Junjie Zhang}, \bibinfo{person}{Zican Dong}, {et~al\mbox{.}}} \bibinfo{year}{2023}\natexlab{}.
\newblock \showarticletitle{A survey of large language models}.
\newblock \bibinfo{journal}{\emph{arXiv preprint arXiv:2303.18223}} (\bibinfo{year}{2023}).
\newblock


\bibitem[Zhao et~al\mbox{.}(2021)]%
        {zhao2021calibrate}
\bibfield{author}{\bibinfo{person}{Zihao Zhao}, \bibinfo{person}{Eric Wallace}, \bibinfo{person}{Shi Feng}, \bibinfo{person}{Dan Klein}, {and} \bibinfo{person}{Sameer Singh}.} \bibinfo{year}{2021}\natexlab{}.
\newblock \showarticletitle{Calibrate before use: Improving few-shot performance of language models}. In \bibinfo{booktitle}{\emph{International Conference on Machine Learning}}. PMLR, \bibinfo{pages}{12697--12706}.
\newblock


\end{thebibliography}

\appendix
\section{Example Saliency Maps for Instruct-GPT}
\begin{figure}
     \centering
     \begin{subfigure}[b]{\textwidth}
         \centering
        \includegraphics[width=0.5\textwidth]{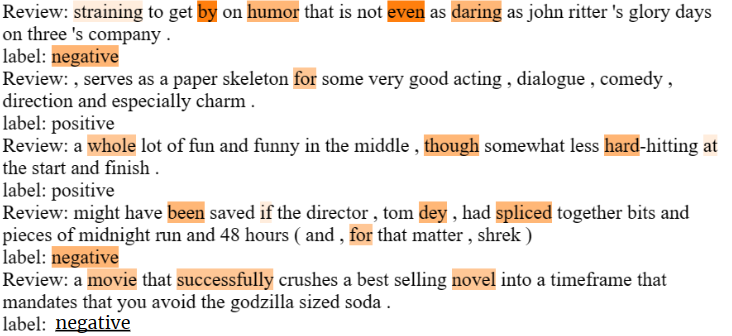}
        \caption{Original prompts (demonstration + test example) used for original demonstration (Instruct-GPT)}
        \label{fig:LIME1}
     \end{subfigure}
     \par\bigskip
     \begin{subfigure}[b]{\textwidth}
        \centering
        \includegraphics[width=0.5\textwidth]{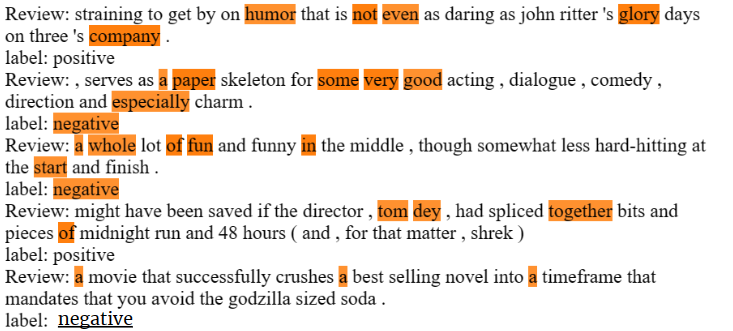}
        \caption{Prompt with label flipping in the demonstration (Instruct-GPT)}
        \label{fig:LIME2}
     \end{subfigure}
     \par\bigskip
     \begin{subfigure}[b]{\textwidth}
        \centering
        \includegraphics[width=0.5\textwidth]{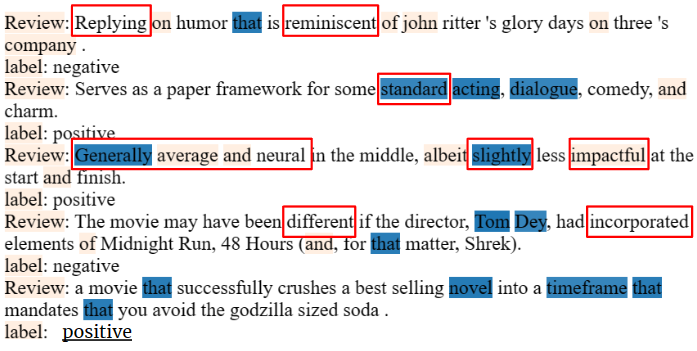}
        \caption{Prompt with input perturbation (neutralization) (Instruct-GPT)}
        \label{fig:LIME3}
     \end{subfigure}
     \par\bigskip
     \begin{subfigure}[b]{\textwidth}
        \centering
        \includegraphics[width=0.5\textwidth]{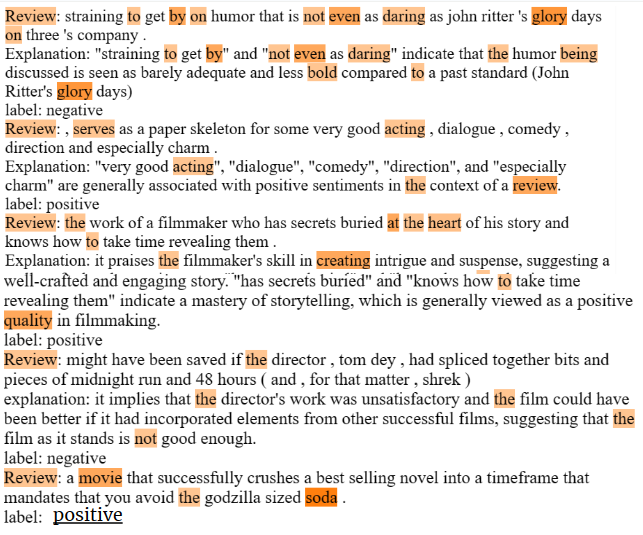}
        \caption{Prompt with complementary explanations in the demonstrations (Instruct-GPT)}
        \label{fig:LIME4}
     \end{subfigure}
        \caption{Full prompts (demonstration + test example) used for original demonstration and three contrastive variants. Tokens are color-coded by saliency scores for generated by LIME. The red box in original and neutralized prompts indicates manually selected sentiment-indicative and sentiment-neutral terms that we used for saliency map comparison.}
        \label{fig:demo_lime}
\end{figure}




\end{document}